\documentclass[sigconf]{acmart}

\usepackage[misc]{ifsym}
\usepackage{subfigure}
\usepackage{subcaption}
\usepackage{multirow}
\usepackage{makecell}
\usepackage[utf8]{inputenc}
\usepackage[most]{tcolorbox}
\usepackage{listings}
\usepackage{algorithm}
\usepackage{algpseudocode}

\copyrightyear{2026}
\acmYear{2026}
\setcopyright{cc}
\setcctype{by-nc-nd}
\acmConference[SIGIR '26]{Proceedings of the 49th International ACM SIGIR Conference on Research and Development in Information Retrieval}{July 20--24, 2026}{Melbourne, VIC, Australia}
\acmBooktitle{Proceedings of the 49th International ACM SIGIR Conference on Research and Development in Information Retrieval (SIGIR '26), July 20--24, 2026, Melbourne, VIC, Australia}
\acmDOI{10.1145/3805712.3809946}
\acmISBN{979-8-4007-2599-9/2026/07}

\begin{document}

\title{M-DaQ: Retrieving Samples with Multilingual Diversity and Quality for Instruction Fine-Tuning Datasets}

\author{Chunguang Zhao}
\affiliation{%
  \institution{Huawei Technologies Ltd.}
  \city{Beijing}
  \country{China}}
\email{zhaochunguang6@huawei.com}

\author{Yilun Liu}
\authornote{Corresponding author.}
\affiliation{%
  \institution{Huawei Technologies Ltd.}
  \city{Beijing}
  \country{China}}
\email{liuyilun3@huawei.com}
  
\author{Pufan Zeng}
\affiliation{%
  \institution{University of Science and Technology of China}
  \city{Hefei}
  \country{China}}
\email{pufanzeng@mail.ustc.edu.cn}

\author{Yuanchang Luo}
\affiliation{%
  \institution{Huawei Technologies Ltd.}
  \city{Xi'an}
  \country{China}}
\email{luoyuanchang1@huawei.com}

\author{Shimin Tao}
\affiliation{%
  \institution{Huawei Technologies Ltd.}
  \city{Beijing}
  \country{China}}
\email{taoshimin@huawei.com}

\author{Minggui He}
\affiliation{%
  \institution{Huawei Technologies Ltd.}
  \city{Beijing}
  \country{China}}
\email{heminggui@huawei.com}

\author{Weibin Meng}
\affiliation{%
  \institution{Huawei Technologies Ltd.}
  \city{Beijing}
  \country{China}}
\email{mwb16@tsinghua.org.cn}

\author{Song Xu}
\affiliation{%
  \institution{University of Science and Technology of China}
  \city{Hefei}
  \country{China}}
\email{xusong@ustc.edu}

\author{Chen Liu}
\affiliation{%
  \institution{Huawei Technologies Ltd.}
  \city{Nanjing}
  \country{China}}
\email{valerie.liu@huawei.com}

\author{Hongxia Ma}
\affiliation{%
  \institution{Huawei Technologies Ltd.}
  \city{Nanjing}
  \country{China}}
\email{mahongxia@huawei.com}

\author{Li Zhang}
\affiliation{%
  \institution{Huawei Technologies Ltd.}
  \city{Shanghai}
  \country{China}}
\email{izzie.zhangli@huawei.com}

\author{Boxing Chen}
\affiliation{%
  \institution{Huawei Technologies Ltd.}
  \city{Montreal}
  \country{Canada}}
\email{boxing.chen@huawei.com}

\author{Daimeng Wei}
\affiliation{%
  \institution{Huawei Technologies Ltd.}
  \city{Beijing}
  \country{China}}
\email{weidaimeng@huawei.com}
  
\renewcommand{\shortauthors}{Zhao et al.}


\begin{abstract}
Multilingual instruction fine-tuning (IFT) empowers large language models to generalize across diverse linguistic and cultural contexts; however, high-quality, systematically curated multilingual IFT datasets remain scarce. To address this gap, we propose \textbf{M-DaQ} (Multilingual Diversity and Quality), a diversity-aware sampling framework that jointly optimizes instruction-response quality and cross-lingual semantic diversity. M-DaQ leverages a fine-tuned Quality Scoring Model alongside a maximal marginal relevance-inspired selection strategy to construct balanced, high-fidelity training data. Furthermore, we present the first systematic investigation of the Superficial Alignment Hypothesis in multilingual settings. Extensive evaluations across 18 languages demonstrate that models trained on M-DaQ-curated data achieve average win rates exceeding 60\% against strong baselines on Alpaca-Eval and MT-Bench. Complementary human evaluations corroborate these gains, highlighting significant improvements in cultural relevance, contextual appropriateness, and instruction-following capability. The code are publicly released to facilitate reproducibility and future research.\footnote{\url{https://github.com/zhaocorey/M-DaQ}}
\end{abstract}



\maketitle

\section{Introduction}
\label{sec:intro}

Instruction Fine-Tuning (IFT) datasets play a pivotal role in enabling Large Language Models (LLMs) to effectively perform general-purpose tasks \cite{alpaca,dolly_15k,vicuna2023}. Empirical studies demonstrate that carefully curated, small-scale datasets (e.g., 1,000 samples \cite{LIMA}) can achieve performance competitive with models trained on significantly larger corpora (e.g., 52,000 samples \cite{alpaca}). Consequently, effective data selection is underpinned by two key factors: (1) preserving linguistic and task diversity \cite{super_natural_instructions,active_instruction_tuning}, and (2) prioritizing samples of high intrinsic quality based on well-defined criteria \cite{from_quantity_to_quality,llm_perplexity_estimation}.

In the context of multilingual LLMs, the construction of the IFT dataset establishes the upper bound of the model's instruction-following capabilities. Currently, LLMs often fail to align with human preferences and cultural nuances in low-resource languages \cite{xllama_100}. For English dataset, Superficial Alignment Hypothesis (SAH) \cite{LIMA} has been proposed, which posits that the IFT stage does not impart new knowledge but merely aligns the model's output with human-preferred response intention and style, while assuming sufficient knowledge acquisition during Continuous Pretraining (CPT). Nevertheless, the validity of SAH in multilingual setting and the influence of the scale of IFT dataset on LLM multilingual capability are not systematically studied yet.

The top three challenges for multilingual IFT datasets are underdeveloped and can be summarized into the following primary challenges:
\textbf{Challenge-1. Absence of extensible IFT sample Quality Scoring Method}: No efficient and language-agnostic scoring method currently exists to effectively evaluate the quality of multilingual data. On the other hand, there are prohibitive costs of manual data collection and curation. Therefore, many ship-flag models are still in shortage of the IFT data samples of low resources. For instance, in the Llama 3 IFT dataset, only 3.01\% of samples are multilingual \cite{llama_3}. 
\textbf{Challenge-2. Study of IFT samples diversity limited on English only}: data diversity has been empirically demonstrated to exert a direct and significant impact on LLM performance during IFT \cite{TACOS}. While prior work has focused predominantly on diversity within monolingual (particularly English) datasets.
\textbf{Challenge-3. Verification of the Superficial Alignment Hypothesis (SAH)}: The applicability of SAH remains unexplored in multilingual settings, specifically regarding how limited instruction data interacts with cross-lingual knowledge transfer. 

To address these challenges, methodologies from Information Retrieval (IR)—specifically diversity-driven retrieval methods \cite{DIVERGE,diversitydrivenqueryrewriting}—offer a promising avenue for IFT dataset construction. Theoretically, "quality" in data selection can be modeled analogously to "relevance" in IR, which are essentially both scoring models. Thus, diversity improvement techniques from IR can be effectively transferred to the domain of data curation.

In summary, this work makes the following \textbf{three contributions}:
\begin{itemize}
    \item Contribution-1. We propose \textbf{M-DaQ, a novel diversity driven samples retrieval method for improving LLM multilinguality}. This method jointly optimizes both Challenge 1, quality scoring, and Challenge 2, diversity driven algorithm, for Multilingual IFT dataset. We will open-source code to facilitate reproducibility and future research.
    \item Contribution-2. We conduct a comprehensive \textbf{multilingual IFT evaluation across 18 languages}. Our results demonstrate that M-DaQ effectively mitigates multilingual challenges arising from the data skewness of multilingual IFT data, and the underrepresentation of low resource culturally specific values.
    \item Contribution-3. We present the first systematic empirical investigation of the challenge-3. \textbf{SAH in multilingual contexts}. In this work, we investigate the validity of SAH in multilingual scenarios by analyzing how variations in the scale and composition of the IFT dataset influence cross-lingual LLM alignment, hence the performance of LLM downstream tasks.
\end{itemize}

\begin{figure}[ht]
\centering
\begin{minipage}{0.9\linewidth}
  \centering
  \centerline{\includegraphics[width=8.5cm]{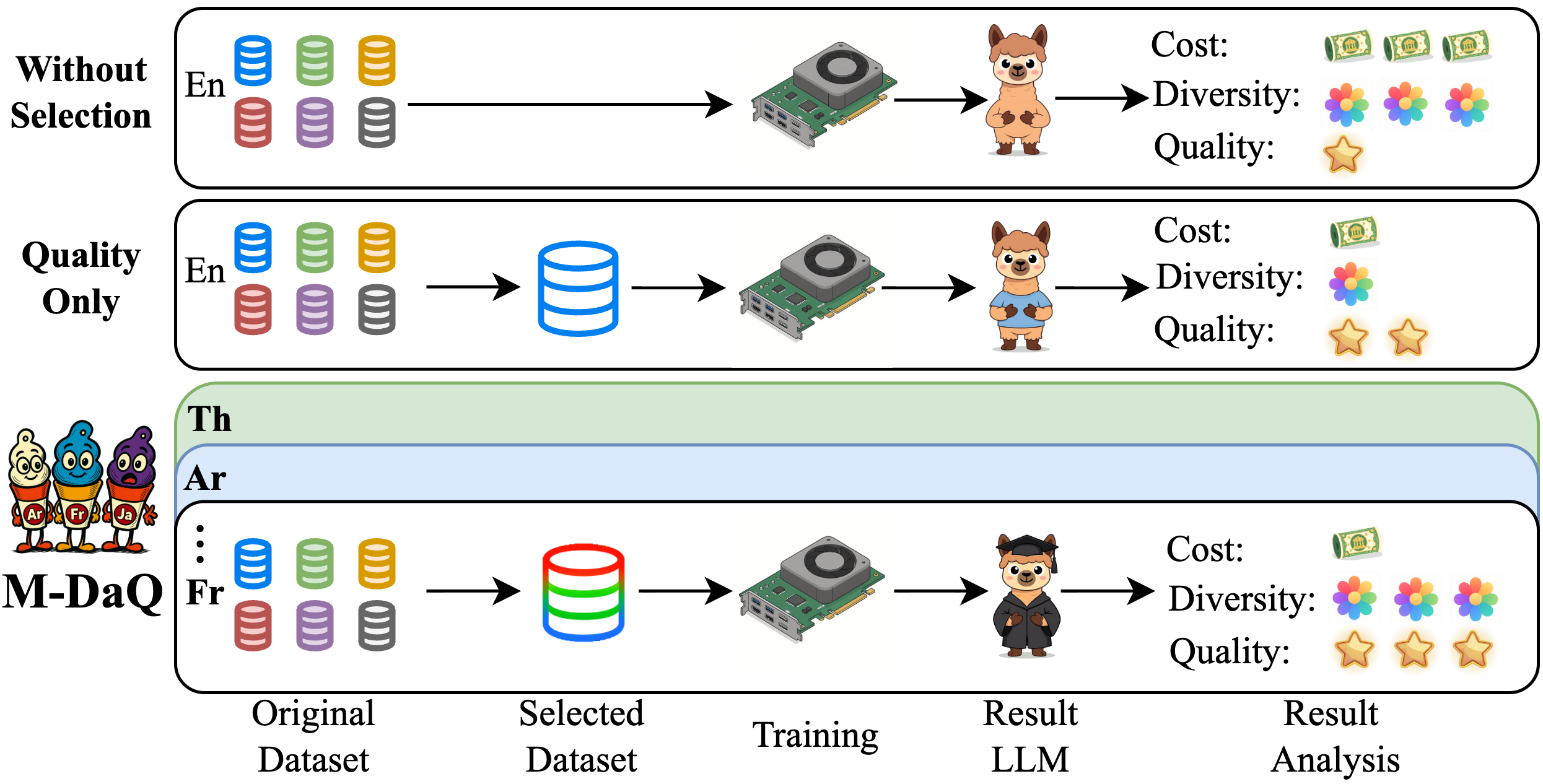}}
  \centerline{(a) M-DaQ data improvement overview}\medskip
\end{minipage}

\begin{minipage}{0.9\linewidth}
  \centering
  \centerline{\includegraphics[width=8.5cm]{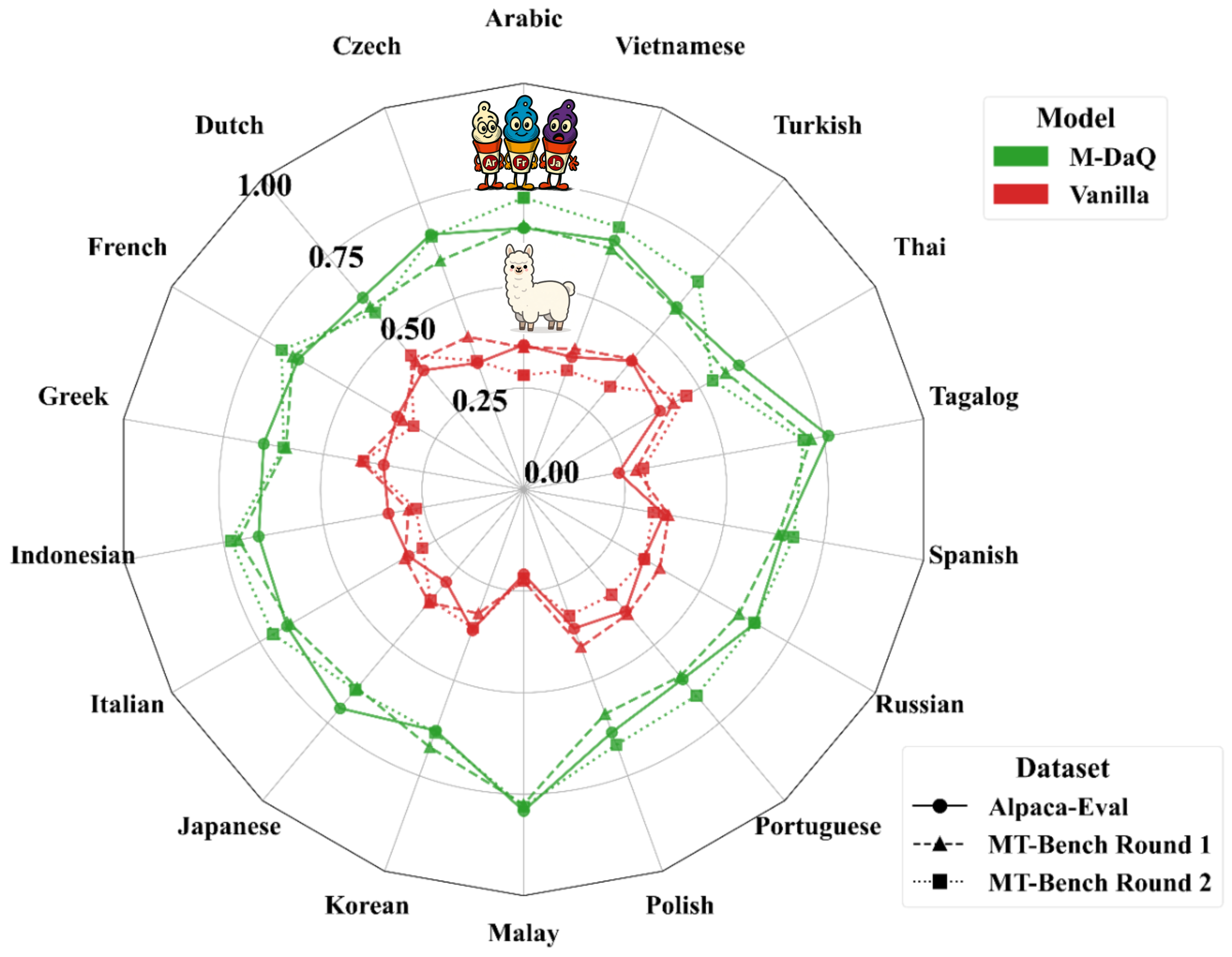}}
  \centerline{(b) M-DaQ Model versus Vanilla Model performance}\medskip
\end{minipage}

\caption{(a) M-DaQ framework comparison: Unlike naive approaches that either use all data (high cost, inconsistent quality) or filter by quality alone (low diversity), M-DaQ's diversity-aware selection achieves optimal balance—reducing computational cost while maintaining high quality and cross-lingual diversity, yielding superior multilingual LLM performance. (b) Cross-lingual performance gains: average win rates of M-DaQ fine-tuned models versus baseline across 18 languages on Alpaca-Eval and MT-Bench. The green lines represent win rates of M-DaQ v.s. vanilla and red lines is vanilla v.s. M-DaQ.}
\Description{Figure (a) presents a comparative schematic illustrating how the M-DaQ pipeline improves upon traditional data curation methods. It highlights three axes of improvement: reduced computational cost through efficient clustering, enhanced response accuracy. Figure (b) displays a horizontal bar chart comparing model performance across eighteen languages. Each language features two adjacent bars representing the baseline and M-DaQ models on two evaluation suites. Across all languages, the M-DaQ bars consistently exceed the fifty percent threshold, demonstrating statistically significant performance improvements.}
\label{fig:hook_image}
\end{figure}

\section{Related Work}

\paragraph{Instruction Fine-Tuning Data Selection.}
IFT data selection has been investigated in several prior studies~\cite{chen2023improvingtranslationfaithfulnesslarge, liu2024makesgooddataalignment, du2023modsmodelorienteddataselection, tree_instruct, car}. For example, Zhao et al.~\cite{long_is_more} propose a remarkably simple yet effective heuristic: selecting the top-$k$ instruction-response pairs with the longest responses from standard datasets, under the intuition that longer responses contain richer learnable signals, are harder to overfit, and encourage models to capture long-range semantic dependencies. Their experiments demonstrate that fine-tuning on merely 1,000 longest examples from Alpaca consistently outperforms more sophisticated selection methods—including manually curated datasets—when evaluated by strong LLM judges. Complementing this heuristic approach, Chen et al.~\cite{alpaca_gasus} introduce an LLM-as-a-judge framework wherein a powerful API model (e.g., ChatGPT) scores each training triplet on dimensions such as accuracy or helpfulness, enabling automatic filtering of low-quality instances. By retaining only examples exceeding a predefined threshold (e.g., score $\geq 4.5$), they construct a compact high-quality subset (~9k samples from Alpaca-52k) that yields superior instruction-following performance with substantially reduced training costs.

However, these methods largely disregard the linguistic and cultural specificities of individual languages. Furthermore, as noted by recent work~\cite{reliable_mulitlingual_llm_judge}, employing general-purpose LLMs as judges for multilingual data quality remains unreliable due to cross-lingual inconsistencies and evaluation biases. To overcome these limitations, we fine-tune an expert model—QSM (Quality Scoring Model) specifically designed for multilingual data quality evaluation, thereby decoupling quality assessment from the biases inherent in generic LLM judges.

\paragraph{Diversity-Aware Selection from Information Retrieval.}
Inspiration for diversity-aware selection can be drawn from the information retrieval (IR) literature. Since the introduction of the Probability Ranking Principle (PRP)~\cite{PRP_book}, retrieval strategies have evolved to balance not only relevance but also result diversity—often formalized as maximizing relevance scores while minimizing pairwise similarity or covariance among retrieved items. A prominent example is the Maximal Marginal Relevance (MMR) algorithm~\cite{MMR}, which explicitly optimizes for both relevance and novelty in result sets. This principle provides a compelling foundation for IFT data selection: by substituting relevance with a quality metric and novelty with linguistic or semantic diversity, MMR offers a theoretically grounded mechanism for curating training sets that are both high-quality and representative across languages and domains.

\section{Method}
\label{sec:method}

\begin{figure}[tbp]
    \centering
    \includegraphics[width=0.75\linewidth]{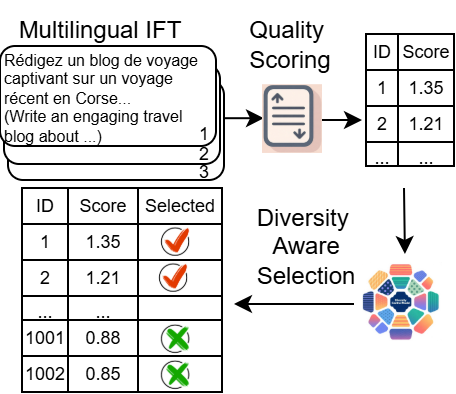}
    \caption{Architecture of the M-DaQ framework. The \textbf{Quality Scoring Model (QSM)} estimates sample quality via triplet loss on expert-aligned preference pairs. The \textbf{Diversity-Aware Selection (DAS)} module applies a MMR-inspired strategy to balance individual sample quality with corpus-level semantic diversity, yielding a compact, representative training subset.}
    \Description{This flowchart outlines the two-stage M-DaQ pipeline. The process begins with a multilingual instruction dataset containing original and expert-revised responses. These samples enter the Quality Scoring Model. The scored dataset then flows into the Diversity-Aware Selection module. The final output is a curated, compact subset optimized for both intrinsic quality and linguistic task coverage, ready for instruction fine-tuning.}
    \label{fig:mdaq_algo}
\end{figure}

\subsection{Quality Scoring Model (QSM)} \label{method_QSM}

We introduce the \textbf{Quality Scoring Model (QSM)}, a fine-tuned model designed to estimate the quality of multilingual IFT samples. The model is trained on a pairwise dataset derived from MIDB \cite{midb}, which comprises approximately 2.3K expert-revised IFT samples per language across 18 languages, originally sourced from the Alpaca dataset. For each instruction $s_i$, we construct a \textbf{positive response} $s_p$ (high-quality, human-revised) and a \textbf{negative response} $s_n$ (low-quality, original). The three most prevalent issues in the negative responses are: (i) content errors inherited from the source (\emph{e.g.}, English) datasets; (ii) artifacts introduced by machine translation; and (iii) insufficient localization, as direct translation often fails to incorporate region-specific cultural context and domain knowledge for low-resource languages. Furthermore, the positive responses are generally more comprehensive and demonstrate deeper reasoning, indicating a clear preference for higher-complexity samples.

The object function is designed to maximize the score for the response more similar to $s_p$, while minimizing the score for the response less similar to $s_n$, effectively learning a relative quality evaluation model. This is formalized via a \textbf{triplet loss} function that operates in the embedding space:

\begin{displaymath}
\mathcal{L}_{\text{triplet}} = \max \left( \| \mathbf{v}_i - \mathbf{v}_p \|_2 - \| \mathbf{v}_i - \mathbf{v}_n \|_2 + \epsilon,\ 0 \right)
\end{displaymath}

where $\mathbf{v}_i$, $\mathbf{v}_p$, and $\mathbf{v}_n$ denote the encoded representations of the instruction, positive response, and negative response, respectively; $\epsilon > 0$ is a margin hyperparameter enforcing separation between positive and negative pairs. This formulation encourages the model to assign higher similarity scores to high-quality (instruction, response) pairs while pushing apart low-quality ones.

\subsection{Diversity-Aware Selection (DAS)} \label{method_DAS}

To ensure linguistic and semantic diversity in the selected multilingual dataset, we propose the \textbf{Diversity-Aware Selection (DAS)} framework---an algorithm that operates across languages without relying on language-specific resources. The algorithm is inspired by MMR criterion. In standard MMR, the candidate maximizing the trade-off between relevance and novelty is iteratively selected. By substituting relevance with QSM scores, Eq. \ref{eq:MMR} formalizes the quality-diversity trade-off. 

\begin{equation} \label{eq:MMR}
    \operatorname{MMR}(d_i) = \lambda \cdot \text{QSM}(d_i) - (1 - \lambda) \cdot \max_{d_j \in \mathcal{S}} \operatorname{Sim}(d_i, d_j)
\end{equation}

To avoid the $\mathcal{O}(n^2)$ greedy optimization in MMR (Eq. \ref{eq:MMR}), we reformulate the selection process into DAS pipeline. The DAS pipeline, formalized in Algorithm~\ref{alg:das}, reformulates the sample selection process by replacing the iterative optimization of Eq.~\ref{eq:MMR} with a computationally efficient two-stage strategy. In Stage~1, we maximize set quality by selecting the top $n_{\text{quality}}$ candidates according to their quality scores, as predicted by the QSM. In Stage~2, we firstly computed embeddings and clusters for all candidates following \cite{ge2024car}, and then scan the remaining candidates and greedily select the highest-quality sample from each previously uncovered cluster until $n_{\text{diversity}}$ diverse samples are collected. By treating cluster membership as a discrete approximation of pairwise similarity, this approach reduces the computational complexity from $\mathcal{O}(n^2)$ to $\mathcal{O}(n \log n)$. Consequently, the continuous trade-off parameter $\lambda$ in the original MMR formulation is realized via the ratio of two interpretable hyperparameters, $n_{\text{quality}}$ and $n_{\text{diversity}}$, which explicitly govern the size of the quality-driven and diversity-driven subsets, respectively. This two-stage procedure constitutes a modified MMR strategy that balances representativeness and redundancy without iterative pairwise comparisons.

\begin{algorithm}[t]
\caption{Diversity-Aware Selection (DAS): Two-stage algorithm for constructing high-quality, semantically diverse multilingual IFT subsets. Stage~1 selects top-$n_{\text{quality}}$ samples by QSM score; Stage~2 greedily augments with diverse samples from previously unseen clusters until $n_{\text{diversity}}$ is reached.}\label{alg:das}

\begin{algorithmic}[1]
\Require IFT Dataset $\mathcal{D}$ with quality scores and pre-computed cluster labels, $n_{\text{quality}}$, $n_{\text{diversity}}$
\Ensure Selected subset $\mathcal{S} \subseteq \mathcal{D}$

\State $\mathcal{D}_{\text{sorted}} \gets \text{Sort}(\mathcal{D}, \text{key} = \text{quality\_score}, \text{descending})$
\State $\mathcal{S} \gets \mathcal{D}_{\text{sorted}}[1:n_{\text{quality}}]$ \Comment{Top-quality samples}
\State $\text{diverse\_set} \gets \{ d.\text{cluster} \mid d \in \mathcal{S} \}$ 

\For{each instruction $d \in \mathcal{D}_{\text{sorted}}[n_{\text{quality}}+1 : ]$} 
    \If{$|\mathcal{S}| \geq n_{\text{quality}}+n_{\text{diversity}}$} 
        \State \textbf{break} 
    \EndIf
    \If{$d.\text{cluster} \notin \text{diverse\_set}$}
        \State $\mathcal{S} \gets \mathcal{S} \cup \{d\}$ \Comment{Diverse samples}
        \State $\text{diverse\_set} \gets \text{diverse\_set} \cup \{d.\text{cluster}\}$
    \EndIf
\EndFor
\end{algorithmic}

\end{algorithm}

\section{Experiment}
\label{sec:experiment}

\subsection{Experimental Setup}

We evaluate our proposed \textbf{M-DaQ} algorithm using the Alpaca-52K dataset, which has been extended to cover 18 languages. In our default configuration, the DAS parameters are set to a quality-to-diversity ratio of $n_{\text{quality}} : n_{\text{diversity}} = 6:1$, determined empirically. We fine-tune the Llama-3-8B base model on the subset selected by M-DaQ, yielding the \textbf{M-DaQ Model}. For baseline comparison, we train a \textbf{Vanilla Model} on the original, unfiltered Alpaca-52K dataset. Both models are trained under identical hyperparameter settings, including a batch size of 256 and a learning rate of $5 \times 10^{-5}$. To ensure training parity, both models undergo exactly three epochs of instruction fine-tuning.

\subsection{Evaluation Protocol}

\textbf{LLM-as-Judge.} We employ a state-of-the-art LLM as an automated judge to conduct pairwise comparisons between the outputs of the M-DaQ and Vanilla models. To mitigate positional bias, we adopt the \textbf{win rate} metric \cite{mt_bench}, calculated as $(\#wins+\#ties)/\#all$, which aggregates results from bidirectional comparisons (i.e., swapping model order across evaluation instances). Evaluations are performed on two widely adopted instruction-following benchmarks: Alpaca-Eval \cite{alpaca_eval} and MT-Bench \cite{mt_bench}. Both benchmarks have been machine-translated and rigorously human-revised to ensure high-quality prompts and references across all 18 languages \cite{midb}.

\textbf{Human Evaluation.} To complement automated assessment and capture nuanced linguistic and cultural dimensions, we conduct a large-scale human evaluation involving seven native-speaking language experts, each with an average of 3.9 years of professional experience in NLP or translation. For each language and benchmark, experts follow a structured multi-stage protocol: random sampling, independent scoring, cross-review, and consensus reconciliation. The evaluation encompasses a total of 900 samples (randomly sampling 50 for each test set and language) and requires approximately 58 person-hours. Consistent with the LLM-as-Judge setup, we report the human-annotated win rate. Experts assess responses across seven dimensions: (1) safety/harmlessness, (2) factual correctness/truthfulness, (3) relevance/helpfulness, (4) completeness/ coherence, (5) readability/fluency, (6) comprehensiveness/diversity, and (7) linguistic naturalness.

\subsection{Results and Analysis}

\begin{table}[htbp]
\caption{Human evaluated win rate of M-DaQ vs. Vanilla Model Performance Across Six Languages on Three Benchmark Tasks}
\label{tab:human_eval}
\centering
\resizebox{\linewidth}{!}{
\begin{tabular}{lccc}
\toprule
Language & Alpaca-Eval & MT-Bench Round 1 & MT-Bench Round 2 \\
\midrule
Japanese & 86\% & 84\% & 80\% \\
Korean   & 94\% & 94\% & 94\% \\
Russian  & 70\% & 86\% & 84\% \\
Portuguese & 80\% & 78\% & 84\% \\
Greek    & 58\% & 76\% & 82\% \\
French   & 80\% & 92\% & 88\% \\ 
\midrule
\textbf{Average} & \textbf{78.00\%} & \textbf{85.00\%} & \textbf{85.33\%} \\
\bottomrule
\end{tabular}
}
\end{table}

\begin{figure}[h]
    \centering
    \includegraphics[width=1.0\linewidth]{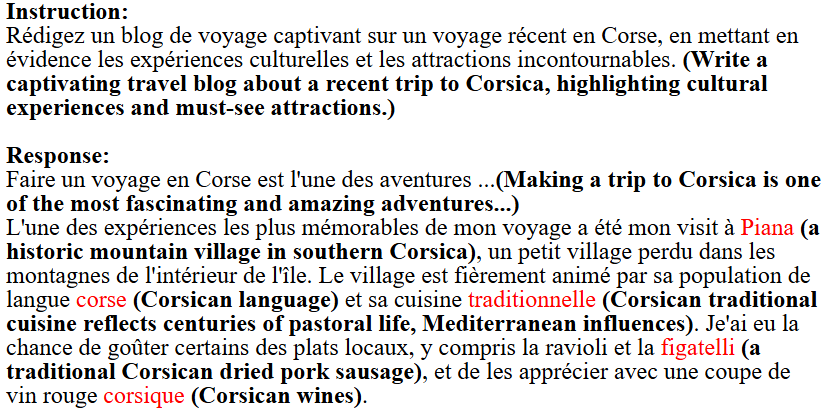}
    \caption{Qualitative example of M-DaQ's cultural localization. For a French travel instruction regarding Corsica, the M-DaQ model incorporates region-specific landmarks (Piana), local linguistic conventions, and traditional cuisine (\emph{figatelli}), demonstrating superior contextual grounding compared to generic multilingual baselines.}
    \Description{This figure presents a side-by-side textual comparison of model outputs for a single culturally specific prompt. The prompt asks a question related to regional customs or local context in a non-English language. The M-DaQ model response incorporates region-specific knowledge, appropriate cultural references, and contextually accurate terminology.}
    \label{fig:culture_example}
\end{figure}
 
\textbf{LLM-as-Judge Results.} Figure~\ref{fig:hook_image}(b) visualizes the automated evaluation across all 18 languages. Key observations include:
\begin{enumerate}
    \item \textbf{Consistent cross-lingual improvement.} M-DaQ model achieves a substantial average win rate of \textbf{60.2\%} on Alpaca-Eval, \textbf{62.6\%} on MT-Bench Round 1, and \textbf{62.9\%} on MT-Bench Round 2, demonstrating robust superiority over the Vanilla baseline across the entire multilingual spectrum.
    \item \textbf{Enhanced gains for low-resource languages.} Performance improvements are particularly pronounced in lower-resource languages (e.g., Tagalog, Malay) relative to higher-resource counterparts (e.g., Dutch, Polish). This indicates that M-DaQ's joint optimization of data quality and diversity effectively mitigates linguistic imbalance and alleviates data scarcity.
\end{enumerate}

\textbf{Human Evaluation Results.} Table~\ref{tab:human_eval} reports the win rates annotated by human for a representative subset of six languages. The average human-evaluated win rate consistently exceeds the automated LLM-as-Judge scores, suggesting that human raters detect qualitative improvements that current discriminative judges may underrepresent. Notably, experts highlighted significant gains in culturally appropriate phrasing, contextual grounding, and domain-specific localization. Figure~\ref{fig:culture_example} illustrates this phenomenon: for a French travel query regarding Corsica, the M-DaQ model generates regionally accurate references and incorporates local linguistic conventions, whereas the baseline produces generic, culturally neutral responses.

\begin{figure}[htb]
    \centering
    \centerline{\includegraphics[width=7cm]{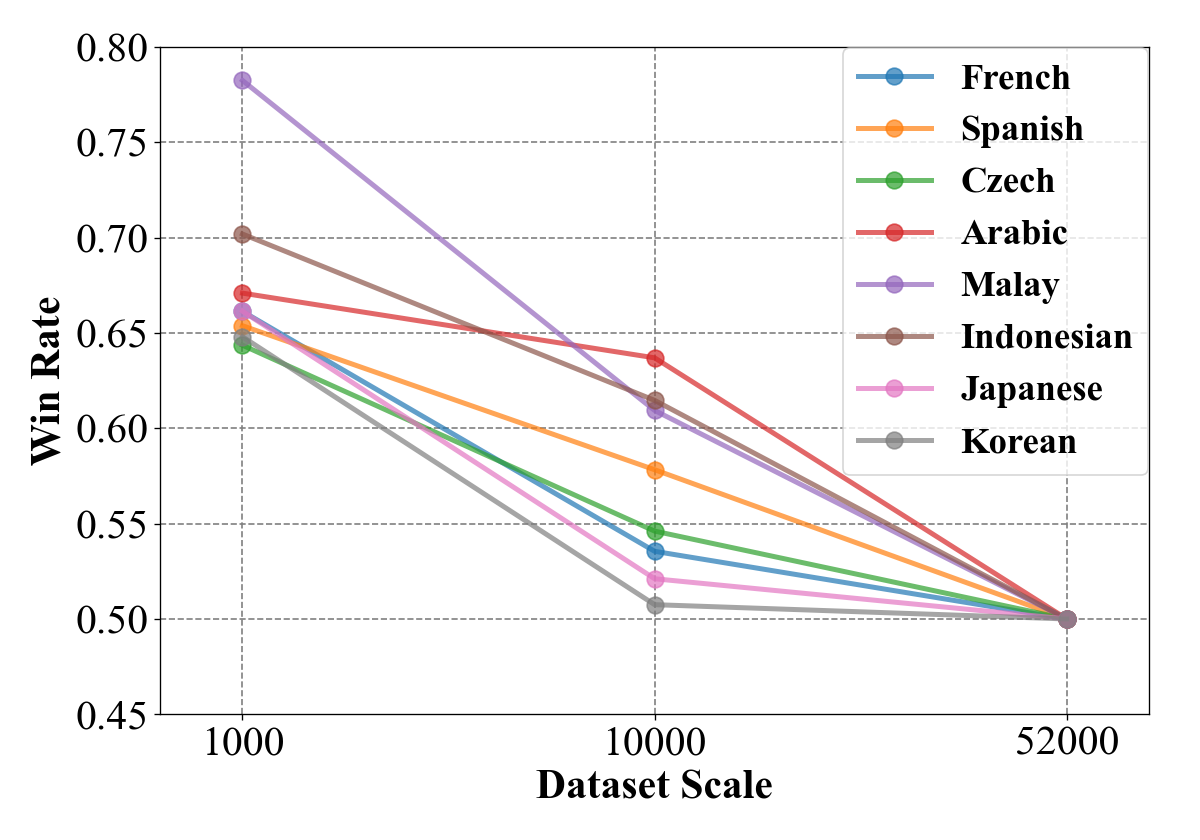}}
    \caption{Validation of the Superficial Alignment Hypothesis (SAH) in multilingual settings: Average win rate versus IFT dataset scale (1K–52K samples) across eight languages. Results indicate diminishing returns beyond small, high-quality subsets, with language-dependent sensitivity to data volume—supporting SAH's applicability while highlighting cross-lingual variation in pretraining readiness.}
    \Description{This line graph plots model performance against instruction fine-tuning dataset size for eight different languages. The horizontal axis represents dataset scale, ranging from one thousand to fifty-two thousand samples. The vertical axis shows the average win rate on evaluation benchmarks. Each language is represented by a distinct colored line.}
    \label{fig:SAH}
\end{figure}

\textbf{Validation of the Superficial Alignment Hypothesis (SAH) in Multilingual Settings.} To examine the cross-lingual applicability of the SAH, we fine-tune Llama-3-8B under three IFT data conditions: 1K and 10K subsets both curated by the full M-DaQ pipeline applied to Alpaca-52K, and the unfiltered Alpaca-52K dataset. Performance is measured by average win rate across Alpaca-Eval and MT-Bench in eight languages (Fig.~\ref{fig:SAH}). Our analysis yields two principal findings:
\begin{enumerate}
    \item \textbf{Diminishing returns with dataset scale.} Expanding the instruction fine-tuning (IFT) dataset from 1K to 10K and 52K samples reduces the average win rate by 10.1\% and 6.2\%, respectively, relative to the 1K high-quality baseline. This confirms that a few thousand carefully curated samples suffice for effective alignment, validating the SAH in multilingual instruction fine-tuning.
    \item \textbf{Language-dependent sensitivity.} The impact of dataset scale varies significantly across languages. For instance, Arabic exhibits a flatter win-rate curve in the 1K–10K range compared to French, indicating lower sensitivity to dataset expansion. We attribute this variation to differences in cross-lingual pretraining readiness, suggesting that the practical utility of the SAH is modulated by a language's representation in the base model.
\end{enumerate}

\section{Conclusion}
\label{sec:conclusion}

In this work, we present \textbf{M-DaQ}, a language-agnostic, diversity-aware sampling framework that jointly optimizes instruction-response quality and cross-lingual semantic diversity for multilingual instruction fine-tuning. By integrating a fine-tuned Quality Scoring Model with a maximal marginal relevance-inspired selection strategy, M-DaQ constructs compact, high-fidelity training subsets that effectively mitigate data scarcity and cultural misalignment—particularly benefiting lower-resource languages. Extensive evaluations across 18 languages demonstrate consistent gains over strong baselines on Alpaca-Eval and MT-Bench, with human assessments confirming notable improvements in cultural relevance, contextual appropriateness, and instruction-following capability. Furthermore, our systematic investigation provides the first empirical validation of the Superficial Alignment Hypothesis in multilingual settings, revealing both its general applicability and language-dependent sensitivity to IFT data scale. We release our code to support reproducible research and future advancements in multilingual LLM alignment.

\bibliographystyle{ACM-Reference-Format}
\bibliography{refs}

\end{document}